\documentclass[letterpaper, 10 pt, conference]{ieeeconf}  

\IEEEoverridecommandlockouts                              

\usepackage{graphics} 
\usepackage{epsfig} 
\usepackage{mathptmx} 
\usepackage{times} 
\usepackage{amsmath} 
\usepackage{amssymb}  
\usepackage{xcolor}
\usepackage{float}
\usepackage{tikz}
\usetikzlibrary{shapes,arrows}
\usepackage{algorithm}
\usepackage[noend]{algpseudocode}
\usepackage{hyperref}
\usepackage{subcaption}
\usepackage[switch]{lineno}
\nolinenumbers
\algblockdefx[NAME]{RepeatParallel}{RepeatUntil}%
[2][\textbf{in Parallel}]{\textbf{Repeat} #1: #2}%
[1][]{\textbf{Until} #1}
\algblockdefx[Name]{Init}{EndInit}%
[1][]{\textbf{Initialize}: #1}%
[1][]{}
\algnewcommand{\Initialize}[1]{%
	\State \textbf{Initialize:}
	\State \hspace*{\algorithmicindent}\parbox[t]{0.8\linewidth}{\raggedright #1}
}
\floatname{algorithm}{Algorithm}

\usepackage{lipsum}
\usepackage{graphicx}
\newcommand\copyrighttext{%
	\footnotesize  This work has been submitted to the IEEE for possible publication. Copyright may be transferred without notice, after which this version may no longer be accessible.}
\newcommand\copyrightnotice
{\begin{tikzpicture}[remember picture,overlay]
		\node[anchor=south,yshift=10pt] at (current page.south) {\fbox{\parbox{\dimexpr\textwidth-\fboxsep-\fboxrule\relax}{\copyrighttext}}};
	\end{tikzpicture}%
}

\title{\LARGE \bf
Multi-FLEX: An Automatic Task Sequence Execution Framework \linebreak to Enable Reactive Motion Planning for Multi-Robot Applications
}

\author{Gaurav Misra$^{1}$, Akihiro Suzumura$^{2}$, Andres Rodriguez Campo$^{3}$, Kautilya Chenna$^{4}$, Sean Bailey$^{5}$, John Drinkard$^{6}$
\thanks{$^{1,}$$^{3, }$$^{5, }$$^{6 }$Gaurav Misra, Andres Rodriguez Campo, Sean Bailey, and John Drinkard  are with OMRON Research Center of America, San Ramon, CA, USA {\tt\small firstname.lastname@omron.com}}%
\thanks{$^{2}$Akihiro Suzumura is with Astroscale Japan Inc., Tokyo, Japan {\tt\small akihiro.suzumura@ieee.org}}%
\thanks{$^{4}$Kautilya Chenna is with Medra Robotics, San Francisco, CA, USA {\tt \small 	kautilya@chenna.me }}}

\begin{document}
\maketitle
\copyrightnotice
\thispagestyle{empty}
\pagestyle{empty}
\begin{abstract}
In this letter, an integrated task planning and reactive motion planning framework termed Multi-FLEX is presented that targets real-world, industrial multi-robot applications.  Reactive motion planning has been attractive for the purposes of collision avoidance, particularly when there are sources of uncertainty and variation. Most industrial applications, though, typically require parts of motion to be at least partially non-reactive in order to achieve functional objectives. Multi-FLEX resolves this dissonance and enables such applications to take advantage of reactive motion planning. The Multi-FLEX framework achieves 1) coordination of motion requests to resolve task-level conflicts and overlaps, 2) incorporation of application-specific task constraints into online motion planning using the new concepts of \emph{task dependency accommodation}, \emph{task decomposition}, and \emph{task bundling}, and 3) online generation of robot trajectories using a custom, online reactive motion planner. This planner combines fast-to-create, sparse dynamic roadmaps (to find a complete path to the goal) with fast-to-execute, short-horizon, online, optimization-based local planning (for collision avoidance and high performance). To demonstrate, we use two six-degree-of-freedom, high-speed industrial robots in a deburring application to show the ability of this approach to not just handle collision avoidance and task variations, but to also achieve industrial applications. 
\end{abstract}
\section{INTRODUCTION}
Robotic automation enables a wide variety of applications in industrial settings such as assembly, welding, palletizing, deburring, and inspection. To realize robotic automation, planning and scheduling of the robot tasks are required. For example: an industrial robot that is used for assembling machine parts may be required to perform tasks such as first picking up a part, then inspecting the part, and then fastening, welding, deburring, or gluing the part in a fixture. Each of these tasks need to be performed in a particular order and may contain multiple individual motions to be executed by the robot. Therefore, the user manually specifies the task sequence and motion assignment of the robot such that the robot completes each task while avoiding collisions with objects in the environment.

When production needs to increase, one option available to users is to increase the total number of robots doing the work. While this can take the form of replicating workcells, adding a robot to an existing workcell, can be attractive due to the lower material cost and smaller footprint.  This creates a ``multi-robot workcell", with the additional need to avoid collisions with other robots in the shared part of the workspace.

While it is possible for users to program multi-robot applications using manually-created interlocking policies~\cite{flordal2007automatic}, such approaches may be too complex for available resources to realize, or may take excessive periods of time for setup, integration, and validation. Interlocking-type approaches are especially challenging when an application has sources of variation, such as alterations in robot task sequence and duration. For example, in a deburring~\cite{ryuh2006robot} application, the part may arrive with inconsistent timing from a conveyor belt and the part size and shape may be different each time, resulting in potential changes to the start time, path followed, and duration of the deburring. In addition, there may be additional inspection steps needed before deburring which can result in task sequence branching, where a piece is first inspected and then may be discarded due to quality issues. There can be potentially hundreds or even thousands of such variations in an application, making manual task sequence and motion assignment cumbersome and prone to errors. 

These programming activities are complex and can result in substantial setup costs, potentially negating the benefits of a multi-robot workcell. In addition, when changes need to be made, these activities need to be repeated, resulting in a solution that has low flexibility.  As a result, multi-robot workcells are currently difficult for users to choose.

Online motion planning has the potential to simplify the complexity of programming multi-robot applications with sources of variations.  With online reactive motion planning, each robot reacts in real-time to changes in the environment (such as the presence of other robots in the shared workspace) to avoid collisions. Current approaches are insufficient, though, because collision avoidance as the sole function can lead to tasks never being completed. For example, robots may never reach their goal (e.g. ``deadlock") or do tasks in an inappropriate sequence.
%
%
\begin{figure*}[t]
	\centering
	 		\vspace{0.15cm}
		\includegraphics[width=0.95\linewidth]{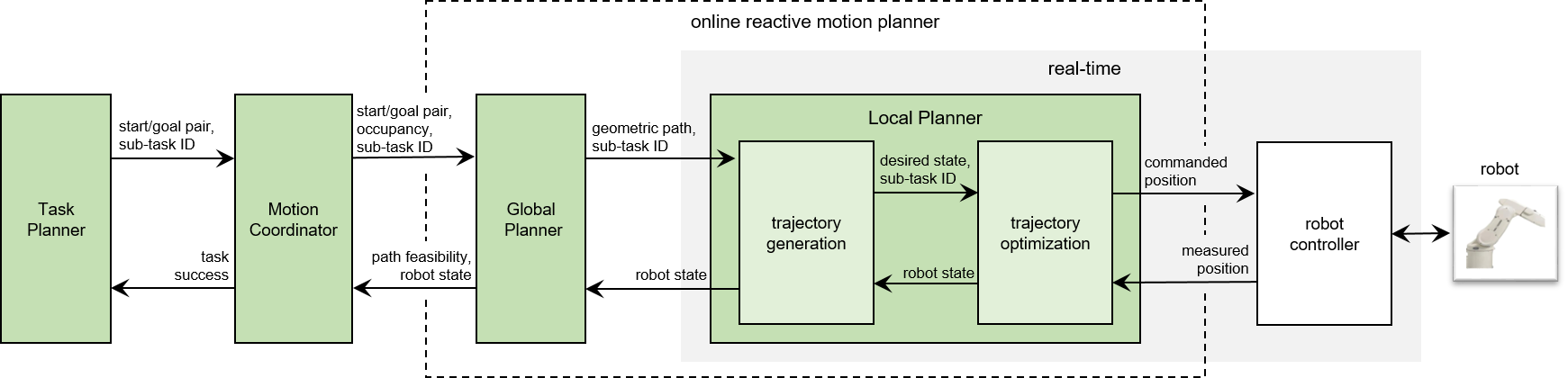}
	\caption{Simplified representation of the Multi-FLEX framework, shown as green blocks. Based on the feedback from the Motion Coordinator, the Task Planner sequentially sends sub-task information to the Motion Coordinator. The Motion Coordinator uses this information to generate robot occupancy information (used for collision avoidance) for the online reactive motion planner. }
	\label{SWInterface}
\end{figure*}

Several approaches have been considered for purely reactive online motion planning for single and multi-robot systems. A local optimization-based planner is proposed in~\cite{bosscher_real-time_2009} for motion planning for multiple six-degree-of-freedom manipulators.
Model predictive control based approaches have also been proposed for multiple SCARA robots in~\cite{al2016hierarchical, he_collaboration_2021}. A hybrid planning approach combining dynamic roadmap-based global planning and optimization-based local planning was proposed in~\cite{volz_predictive_2019} and verified on dual-arm robot manipulators. 

For industrial applications, runtime modifications of attributes of certain types of tasks (like the tool tip path during a welding task) would lead to application failure, so it is desirable for the robot to not always be reactive. Therefore, purely reactive motion planners such as those mentioned in the previous paragraph are insufficient to realize application success.

To fully realize online reactive motion planning, tasks also need to be decomposed, sequenced, and then associated with application-specific constraints before a suitable robot motion plan can be generated. For applications with sources of variation, these constraints cannot be known \emph{a priori} and cannot be fixed due to changes in the environment during runtime. 

Recent research efforts have also looked at individual aspects of task sequencing and coordination for multi-robot application. In~\cite{behrens2020simultaneous}, a simultaneous task assignment and motion scheduling approach based on constraint programming is introduced. In~\cite{touzani2022efficient}, an iterative algorithm is proposed for task sequencing in multi-robot applications. These approaches have focused on improving application cycle time and robot-robot collision avoidance. However, sources of variations in the tasks and accommodation of application-specific constraints (such as segments of motion that must be non-reactive) have not been considered to the best of our knowledge, and both are typical characteristics of industrial applications.

The main contributions presented in this letter are as follows: 1) We present an online task execution approach where tasks can be decomposed into sub-tasks with application-specific constraints and objectives. In addition, we introduce the concept of `bundling', where certain tasks and sub-tasks can be grouped and assigned to one specific robot during runtime; 2) We present a method for how such task assignment can be easily interfaced with an online motion planner, such as MPC, and show a particular implementation of this using a custom, online reactive motion planner; 3) We present a framework for motion coordination between multiple robots to accommodate task-based goal conflicts and overlaps; 4) We demonstrate our approach on an application with sources of variation with multiple robots.

The letter is structured as follows. Section~\ref{systemdesign} describes the system architecture and hardware components used, Section~\ref{TP} describes the Task Planner along with the concepts of task decomposition and bundling, Section~\ref{MC} describes the motion coordination approach proposed to handle goal overlaps and conflict, Section~\ref{OMP} describes the custom, online reactive motion planner, and Section~\ref{Results} presents the deburring application and the experimental implementation with two robots. 

\section{Multi-FLEX System Design}\label{systemdesign}
Multi-robot applications require the decomposition of high-level tasks (e.g. picking, placing, assembling) that form the building blocks of an application, while taking into account constraints at the motion level. These motion-level constraints include collision avoidance between the robot and other robots, the robot and static objects, kinematic constraints, and joint limits. The result is motion profiles for each robot that do not violate constraints at task and motion level during task execution. 

An integrated framework called Multi-FLEX is presented in this letter which includes several functional components, including a Task Planner, a Motion Coordinator, and a custom online reactive motion planner comprising of a Global Planner and a Local Planner as described in~Fig. \ref{SWInterface}. 

\section{Multi-FLEX Task Planner}\label{TP}
In order to generate low-level inputs for robot motion planning, an application must be first decomposed into tasks with associated constraints, objectives, and properties.

We present a Task Planner framework that enables the specification of an entire application into a number of tasks (and associated sub-tasks) that may be sequential or parallel as illustrated in Fig.~\ref{dependencygraph}. Each task is parameterized by
	\begin{itemize}
		\item	The dependency of completion of any other task (see Section~\ref{taskdependency}).
		\item One or more sub-tasks (see Section~\ref{task_decomp}).
			\item The option to declare if a sub-task is part of a bundle (see Section~\ref{task_bundling}).
		\item The robot(s) allowed to perform the task.
	\end{itemize}
\subsection{Task dependency accommodation}~\label{taskdependency}
\begin{figure}[h]
		\centering
		\includegraphics[scale =0.175]{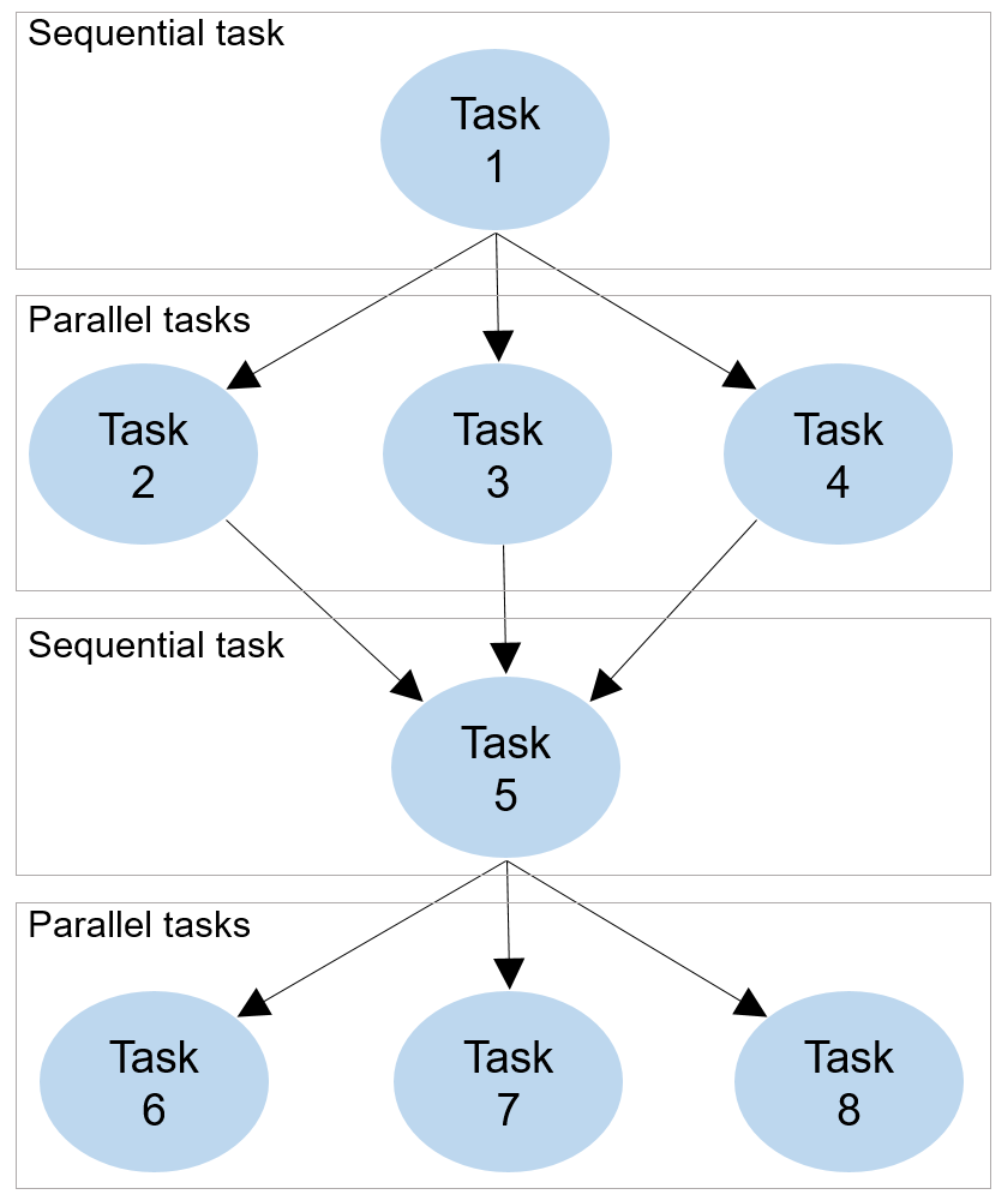}
		\caption{An example of a task dependency graph. Multi-FLEX can dynamically assign parallel tasks to robots as they become available, but waits until sequential tasks are completed before moving on.}
		\label{dependencygraph}
	\end{figure}

	The graph shown in Fig.~\ref{dependencygraph} helps maintain the order of tasks to be performed and task dependencies for one or more robots. For example, in the graph, Task 5 depends on the completion of Task 2, 3, and 4. This dependency constrains the Task Planner from sending a request of Task 5 to the Motion Coordinator until Tasks 2, 3, and 4 are complete. Note that it is possible to define multiple dependencies as input depending on the application constraints. 

	The graph is traversed by using breadth-first search to find the next task of the robot.
	Each task can be available to multiple robots that are identified by the robot number and ID, and the Task Planner automatically assigns priority to the robots and does the task allocation. As a result, the Task Planner restricts execution of the current task unless all of its dependent tasks have been completed successfully. In addition, within the specified task, sub-tasks are created and bundled or not (see Section~\ref{task_bundling}).
\subsection{Task decomposition}\label{task_decomp}
Tasks can be decomposed into sub-tasks with different properties, constraints, and objectives. Previous research on task decomposition and task allocation has considered decomposition of abstract tasks into subtasks and applied these concepts to soccer robot systems~\cite{chen2010research}. 
\begin{figure}[h]
	\centering
	\includegraphics[scale =0.24]{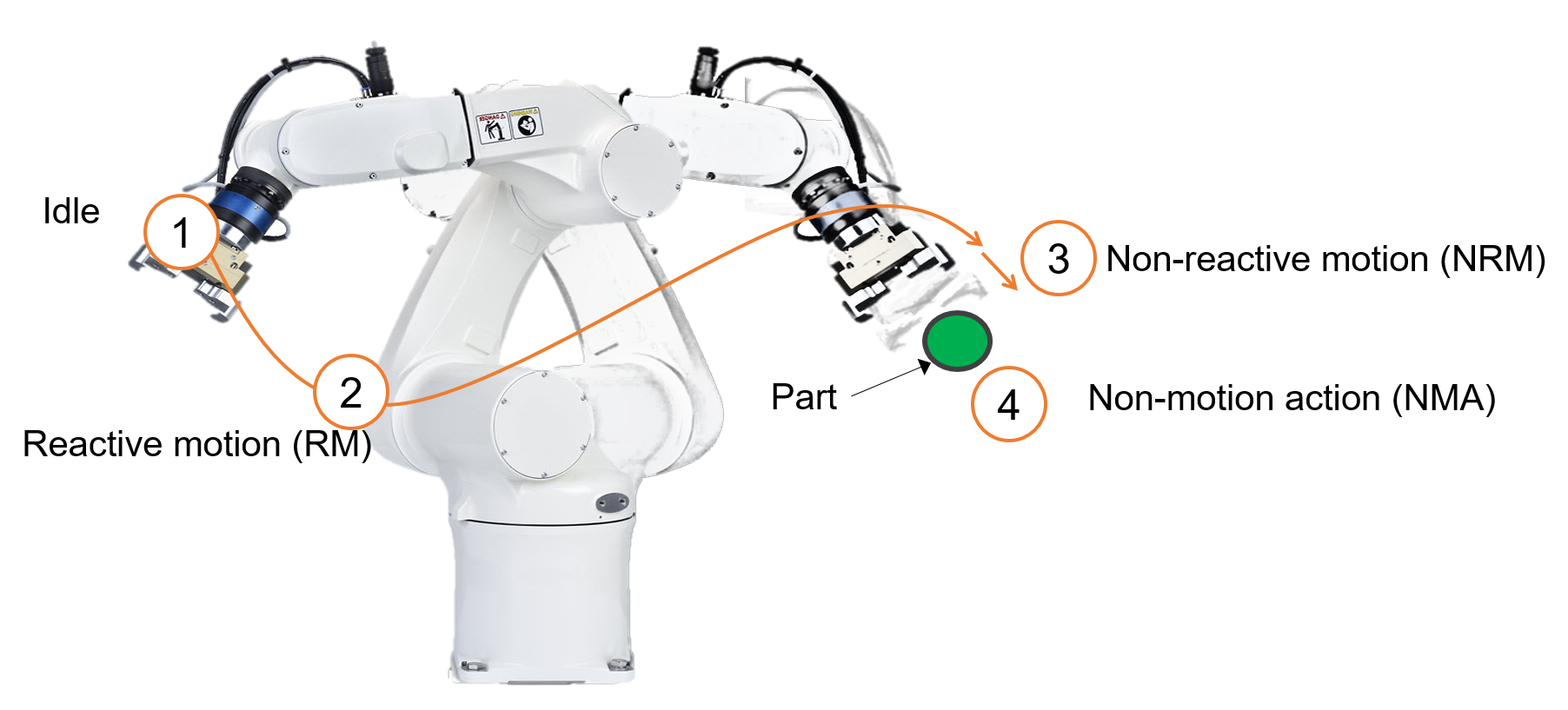}
	\caption{Categories of sub-task types during robot task execution. 1) Idle (robot maintains its pose for a given duration) 2) Reactive motion, or RM (online replanning \emph{with} reactive collision avoidance, e.g., the final approach before picking a part) 3)  Non-reactive motion, or NRM (path/speed constrained motion \emph{without} reactive collision avoidance), and 4) Non-motion action, or NMA (e.g. gripper open/close to grasp the part shown in the figure).}
	\label{tp_motion_category}
\end{figure}

In this work, we present a decomposition framework based on making it possible for low-level online reactive motion planning to handle high-level application-specific constraints.

For example, consider a gluing application where the robot applies a glue-bead. During the deposition of the glue bead, the robot moves the dispensing end-effector with specific path and speed constraints such that the bead is appropriately applied to the part. This type of motion typically cannot be reactive, as any reactive avoidance behavior by the robot will result in part damage and application failure. In addition, there are many applications like this where the system is made to work for only one set of system parameters (e.g. with a certain robot speed for fluid application or a specific dwell time for heat staking), which even eliminate the options of slowing or just stopping the robot.

 The decomposition illustrated in Fig.~\ref{tp_motion_category} divides a task into multiple sub-tasks based on reactivity constraints. The sub-task types are: 
\subsubsection{Idle}
The first category is Idle, where the robot tries to maintain its position but is reactive. This category supports sub-tasks where the robot is required to wait for a given duration at the current robot location before or after beginning a task, due to application-specific constraints. This sub-task type is specified with an ID called ``Idle'' and a predefined time duration. 
\subsubsection{RM}
The second category is reactive motion, where a robot moves from a start position to a given goal position and the Multi-FLEX framework automatically re-plans trajectories during the motion to avoid collisions with other objects or robots in the environment. This sub-task type is specified with an ID called ``RM" (Reactive Motion), a start joint configuration, and a goal joint configuration or an end-effector position and orientation. 
\subsubsection{NRM}
The third category is non-reactive motion, where a robot moves from a start position to a goal position with a pre-defined path and/or velocity \emph{without} reacting to other objects or robots in the environment. Even though these motions are non-reactive, the Multi-FLEX framework automatically handles collision avoidance by first assigning priority to the robot undergoing the non-reactive motion, then computing the volume swept by the robot body during this motion and finally reserving this volume~\cite{behrens2020simultaneous} such that no other object or robot can enter until the robot completes the sub-task. This sub-task type is specified with an ID called ``NRM" (Non-Reactive Motion) and the specified motion. 
\subsubsection{NMA}
The fourth category is non-motion action, where the robot body does not move, but the end-effector may execute some action, such as the opening or closing of the gripper. The volume swept during the end-effector action lies within the shape specified for the end-effector and thus, only the current robot occupancy is reserved. For cases where the volume swept during end-effector action exceeds the specified end-effector shape, the entire swept volume is reserved and the sub-task is treated as an NRM. This sub-task type is specified with an ID called ``NMA" (Non-Motion Action) and required parameters such as the time duration for gripper acquisition or release. 
\subsection{Task Bundling}\label{task_bundling}
The task decomposition framework described above allows for specified sub-tasks to be ‘bundled’ together, if desired, such that the robot always executes these sub-tasks in a given order. For example, consider an ``approach, pick, depart" scenario: the robot end-effector follows a linear path to approach an object, the robot grasps the object, and then the robot departs along the same linear path. These three sub-tasks need to be completed by the same robot and must happen in that sequence. In addition, this sequence must happen without any interference, e.g. from another robot, even if that other robot has higher priority, since an interference during this sequence could lead to application failure. 

To prevent such events from occurring, the framework allows for sub-task bundling.
\begin{figure}[h]
	\centering
	\includegraphics[scale =0.22]{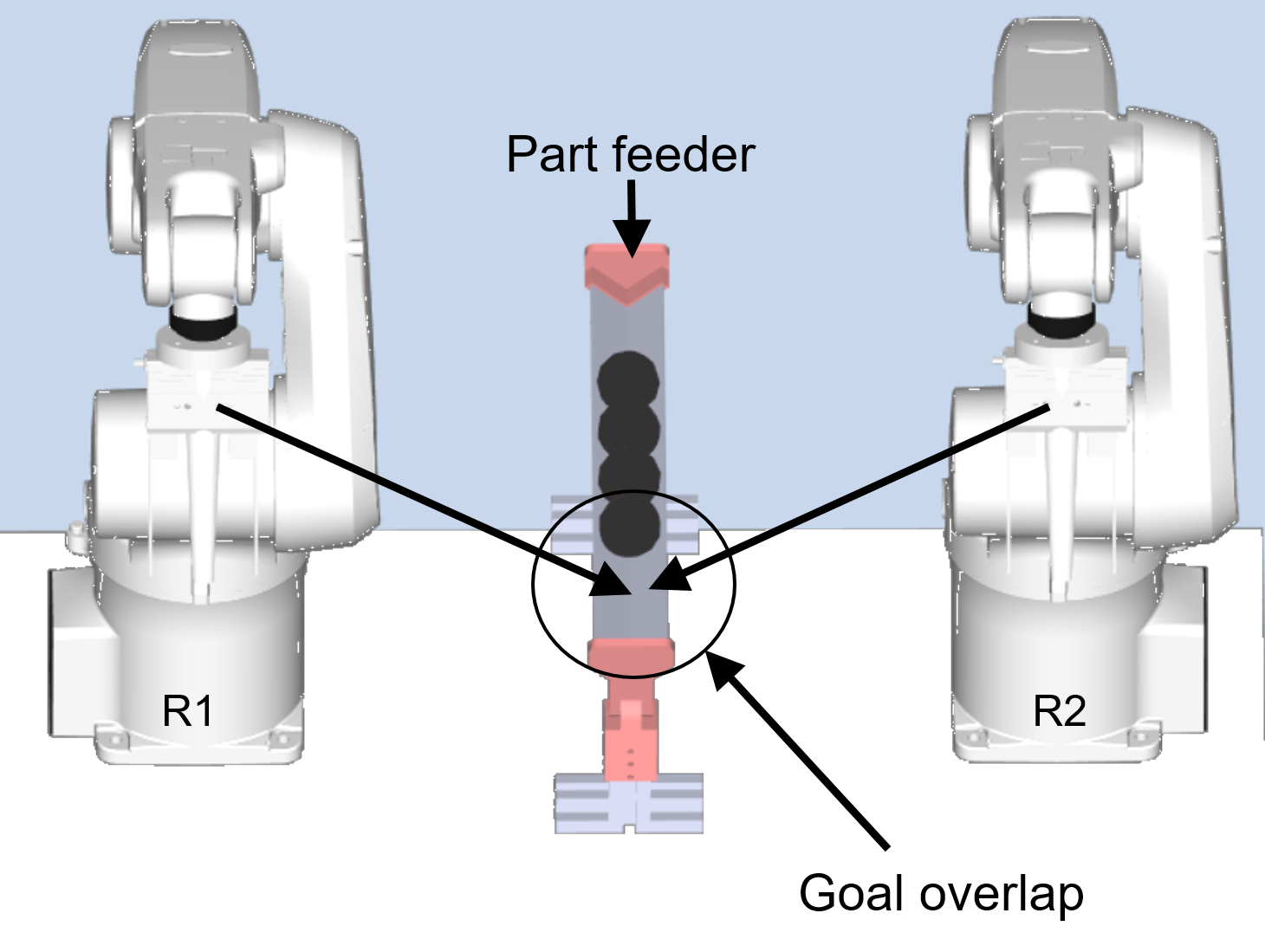}
	\caption{Illustration of goal overlap between two robots, labeled R1 and R2. Both robots are assigned to pick an object from the part feeder (in the center, between the robots), thus creating a goal overlap. The Motion Coordinator de-conflicts the goal overlap situation by assigning priority and manipulating the occupancy information artificially.}
	\label{goalconflict}
\end{figure}
\section{Multi-FLEX Motion Coordinator} \label{MC}
The Motion Coordinator handles goal overlap situations. During online task and motion planning, it is possible that multiple robots are assigned either a) goals (RM) or b) motions or paths (NRM) that cause robot pose overlap, which we call ``goal overlap".

 For example, in Fig.~\ref{goalconflict}, R1 is assigned the task to pick an object from the feeder and the goal pose associated with this task overlaps with the assigned goal pose of the other robot, R2. A reactive motion planner can avoid collision in such situations, but may result in deadlock, where neither robot reaches the goal. In addition, in some situations, the robots oscillate continuously - making some progress toward the goal but then reactively retreating before reaching the goal. 

 As a result, if path planning for the robots happens when there is goal or (pre-defined) path overlap, it is possible that these robots will deadlock and be unable to complete their respective tasks. Obviously, these situations can bring productivity to a stop.
 
\begin{figure}[h]
	\centering
	\includegraphics[scale =0.22]{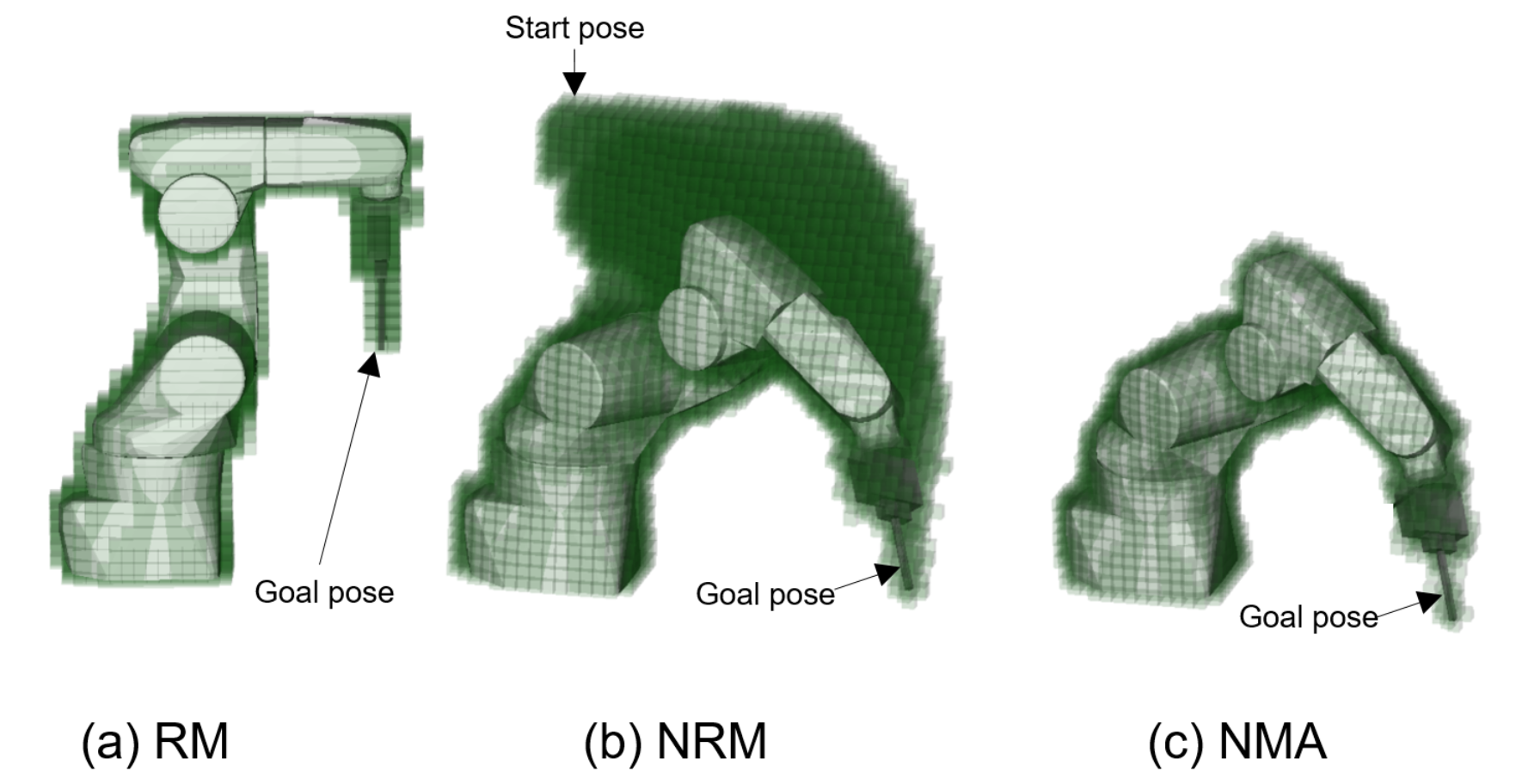}
	\caption{Illustration of voxel-based occupancy reservation for RM, NRM, and NMA, shown in green. In Figures (a) and (c), the robot occupancy is reserved only at the goal pose. In Figure (b), the entire point-to-point path is reserved. So, specific paths/motions required by the application (NRMs) are protected from interruption, while other motions (RMs) are left to be dynamically updated during the motion.}	
	\label{MCReservation}
\end{figure}
The Motion Coordinator handles these overlap situations by manipulating the occupancy information artificially.
 \begin{figure*}[h!]
 	\vspace{0.15cm}
	\centering
	\includegraphics[width =0.85\linewidth]{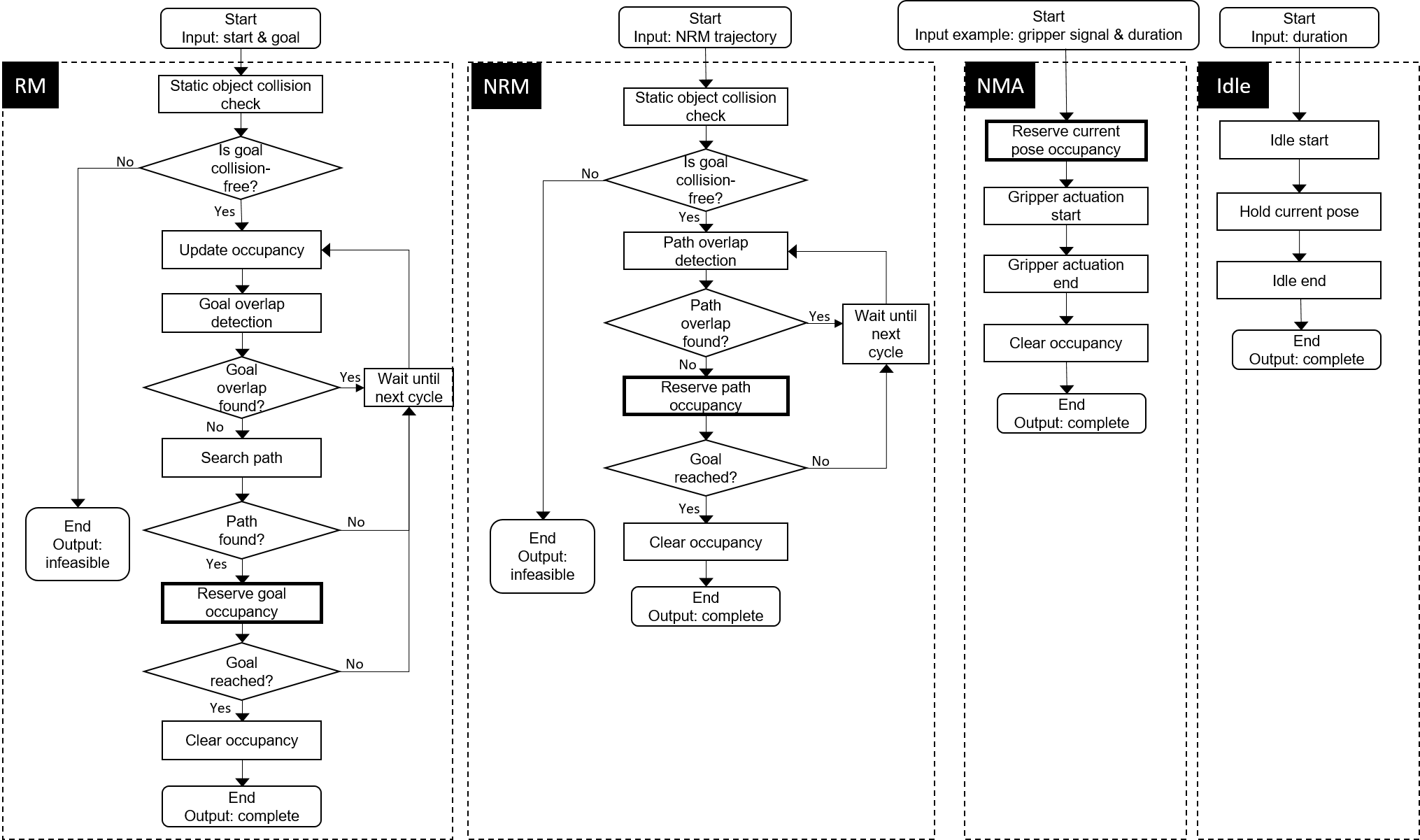}
	\caption{Flowcharts for RM, NRM, NMA, and Idle sub-tasks. For RM, goal overlap detection and path search occurs before goal reservation. For NRM, the entire occupancy for the motion is reserved after overlap detection is completed. For NMA, reservation is done at the current pose. For Idle, no reservation is required.   }	
	\label{MCalgorithm_RM}
\end{figure*}

In such cases, the high priority robot’s goal pose or the swept volume in case of a predefined path is registered as occupied for all other robots. This ensures that the lower priority robots cannot find a feasible path to the goal and need to wait until the high priority robot successfully completes its task.

 The output of the Motion Coordinator is spatial constraints (i.e., regions in the robot workspace/configuration space through which paths are allowed), based on current robot occupancy, robot goals, and the assigned sub-task. These spatial constraints are used downstream by the Global Planner. During each planning cycle of the Global Planner, the occupancy of the robots are updated and goal overlap is checked with respect to other robots using a voxelized representation of the robots.
  
 Fig.~\ref{MCalgorithm_RM} shows a flowchart of how the goal overlap detection and reservation features of the Motion Coordinator are used during Idle, RM, NRM, and NMA sub-tasks. In all cases, if goal overlap is found, the robot enters into a `Wait' state at its current configuration until the next planning cycle. The occupancy reservations for all sub-task types are described in Fig.~\ref{MCReservation}. For Idle, no reservations occur. For RM, for the case where there is no goal overlap, the path search is conducted and the goal pose is reserved if the path search is feasible. For NRM, the entire point-to-point path (referenced in Fig.~\ref{MCReservation}) is reserved. For NMA, the already-occupied goal pose is reserved. The NRM category also includes large end-effector cases where the gripper opening/closing requires volume reservation; once the robot has reached the goal, the occupancy is cleared. 

\section{Multi-FLEX Online Motion Planner}\label{OMP}
The Multi-FLEX online motion planner is a custom hybrid online planner that consists of a Global Planner (for geometric path planning) and a Local Planner (for trajectory generation and local trajectory optimization) to generate collision-free motion for multiple robots. 

\subsection{Global Planner}
  In the Multi-FLEX framework, every robot is associated with an independent and unique Global Planner. 
The Global Planner is based on the concept of dynamic roadmaps~\cite{leven2002framework}. Due to their multi-query nature, dynamic roadmaps are well suited for industrial applications with potentially constantly changing and unique start/goal configurations. 

The Global Planner consists of an offline component and an online component. 
The offline component is a roadmap generator. The roadmap generator creates a collision-free graph $G(V,E)$ and a voxel-graph map.
The online component is made up of a roadmap updater and a dynamic path planner. 
During runtime, the roadmap updater identifies collision-free vertices and edges and then updates the graph. The roadmap updater utilizes the voxel-graph map and runtime-updated workspace voxel occupancy to update the graph.

The presence of other robots is considered during offline roadmap generation, and the dynamic occupancy of other robots is used during runtime to update the individual roadmaps prior to path search.

 The online dynamic path planner uses a typical $\mathrm{A}^{*}$ algorithm to conduct path search. The output of the global planner is collision-free geometric paths for each robot.
  \begin{figure*}[h!]
  	\vspace{0.15cm}
 	\centering
 	\includegraphics[width=0.65\linewidth]{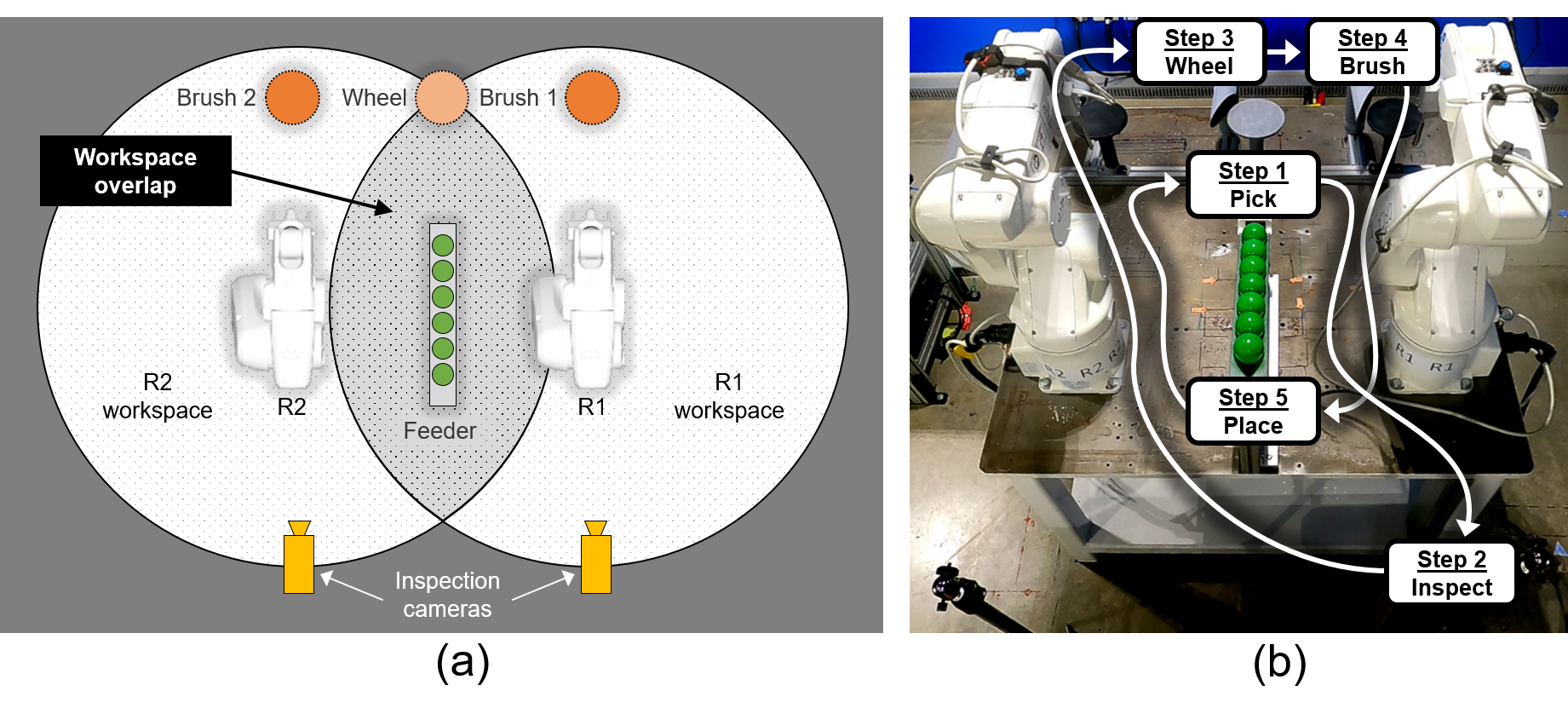}
 	\caption{Top down view  of the deburring workcell, with two robots, labeled R1 and R2. Figure a illustrates the overlapping workspace region in center region, shown in dark grey.  Figure b illustrates the five tasks within a single part deburring application. This workcell is based on a deburring application shown in~{https://youtu.be/0hzcKwHoJdw}. The shared workspace contains areas critical to the application (picking and placing locations, and material processing, i.e. center wheel) as well as transit areas (e.g. from the inspection camera to the deburring equipment) and is thus heavily trafficked.}
 	\label{deburringstation}
 \end{figure*}
 
\subsection{Local Planner}
The Local Planner takes geometric paths as inputs for one or more robots and generates time-parameterized collision-free motion as output. The output generation occurs in two sequential steps that are described below.
\subsubsection{Trajectory generation}
To generate a feasible trajectory from a geometric path, a linear segment with a parabolic blend-type trajectory is constructed with joint synchronization~\cite{biagiotti2008trajectory}. This type of trajectory generation is analytical and works well with rest-to-rest motion.

 To handle cases with non-zero initial velocity, a heuristic algorithm is used to a) compute robot velocity at intermediate waypoints and b) solve a two-point boundary value problem between a pair of waypoints, based on the approach in~\cite{kunz2014probabilistically}. This Multi-FLEX approach generates a joint angle and velocity profile that passes through all geometric path waypoints. 
 
 
 This approach is sub-optimal in terms of time or energy, but is fast to compute and enables trajectory generation in real-time.

  The generated trajectory is sent incrementally to the next step in the Local Planner (trajectory optimization) at a given sampling period (e.g. 16 ms). 
\subsubsection{Trajectory optimization}
Trajectory optimization uses the desired angle and velocity to generate collision-free motion. A constrained optimization problem is solved to generate the commanded joint positions while ensuring collision avoidance constraints, bounds on joint angles, velocity, and acceleration are satisfied. This is formulated as a convex quadratic program (QP) that minimizes an objective function. The deviation between the current joint configuration and the desired waypoint over a single step is used during formulation of the objective function. This approach is similar to a one-step model predictive control problem~\cite{killpack2016model}. 

The decision variables for the QP include the joint velocity and acceleration, denoted as $u = \begin{bmatrix} \dot{q}[k+1], \ddot{q}[k+1]\end{bmatrix} \in \mathbb{R}^{6}\times \mathbb{R}^{6}$. The collision avoidance constraints in the QP are expressed as linear inequalities in terms of the joint velocity. The QP shown in Eq.~\ref{QP} is solved at each Local Planner sampling period.
\begin{equation}
	\begin{aligned}
		& \underset{u}{\text{minimize}}
		& &u^{\mathrm{T}}\mathrm{H} u +
	\begin{bmatrix} 0_{1\times 6} \\ \mathrm{K_{p}}\Delta q[k]+\mathrm{K_{d}\Delta\dot{q}[k]}    \end{bmatrix}^{\mathrm{T}} u\\
		& \text{subject to}
		& & g(u)\leq 0 \\
		&&& c(u) = 0
	\end{aligned}
\label{QP}
\end{equation}
where $\mathrm{H} = diag(\epsilon_{1\times 6}, 1_{1\times 6}) \in \mathbb{R}^{12\times 12}$; $\epsilon << 1 $; and $\mathrm{K_{p}}, \mathrm{K_{d}} \in \mathbb{R}^{6}$ are gains for a PD type controller that tracks the desired joint position and velocity. In addition,  $\Delta q[k] = q_{d}-q[k]$ and $\Delta \dot{q}[k] = \dot{q}_{d}-\dot{q}[k]$ are the errors  in joint position and velocity, where $q_d$ and $\dot{q}_{d}$ are the next desired joint position and velocity.

 The inequality constraint $g(u)\leq 0$ includes collision avoidance constraints and joint limit constraints. The collision avoidance constraint requires distance information between ``closest point pairs" on the robot and other objects in the environment. The robot links are modeled using capsules and other objects in the environment are modeled using swept-sphere primitives. The capsule to swept-sphere distance computations are fast to compute, making it suitable for real-time implementation. 
 
 The collision avoidance constraint for a closest point pair is based on~\cite{bosscher_real-time_2009} and is expressed as
\begin{align}
	\hat{d}J(q[k])\dot{q}[k+1]\leq \hat{d}v_{r}-\alpha \log({\frac{d_{react}-\|d\|}{d_{react}-d_{eq}}})
\end{align}
where $d \in \mathbb{R}^{3}$ is the vector between the closest point pair on the robot and object, $\hat{d} = \frac{d}{\|d\|}$ is the unit vector, $J(q[k]) \in \mathbb{R}^{3\times 6}$ is the closest point translational Jacobian, $v_{r} \in \mathbb{R}^{3}$ is the velocity of the object, $\alpha$ is a tuning parameter, $d_{react}$ is the distance below which the collision avoidance constraint is active, and $d_{eq}<d_{react}$ is the equilibrium distance. 
When the closest distance $\|d\|$ is equal to the equilibrium distance, the allowable relative approach speed becomes zero or negative. Therefore, each closest point pair maintains at least this equilibrium distance, ensuring that collisions do not occur.

 If the NRM or NMA sub-task is active, the collision avoidance constraints are deactivated to ensure the robot strictly follows the desired trajectory without any reactive avoidance. 
 
The equality constraints $c(u)= 0$ are used to impose kinematic constraints on the motion.

In the end, the computed collision-free joint poses from the Local Planner are sent to the robot controller for execution. 
\section{Results}\label{Results}
\subsection{Application}
The functionality and performance of the proposed Multi-FLEX system are demonstrated experimentally using a deburring application with two robots. Deburring is the process of removing small imperfections known as burrs from machined metal products. Multi-robot deburring applications often have overlapping robot workspaces. They can also have long duration, multi-step processes with multiple sources of variation such as changes in part size, branching, and part rejection. Online reactive motion planning is suitable for such type of application characteristics as the motion can be re-planned based on changes in the environment and can increase productivity in applications with these types of characteristics. 

A deburring application workcell with two robots was set up as shown in Fig.~\ref{deburringstation}. The workcell consists of two OMRON Viper650 six degree-of-freedom industrial robots with two-finger grippers, resulting in a combined reach of 790 mm. The centers of the robot bases are separated by 820 mm, which results in significant operation workspace overlap. 
\begin{figure}[h]
	\centering
	\includegraphics[width=0.125\textwidth]{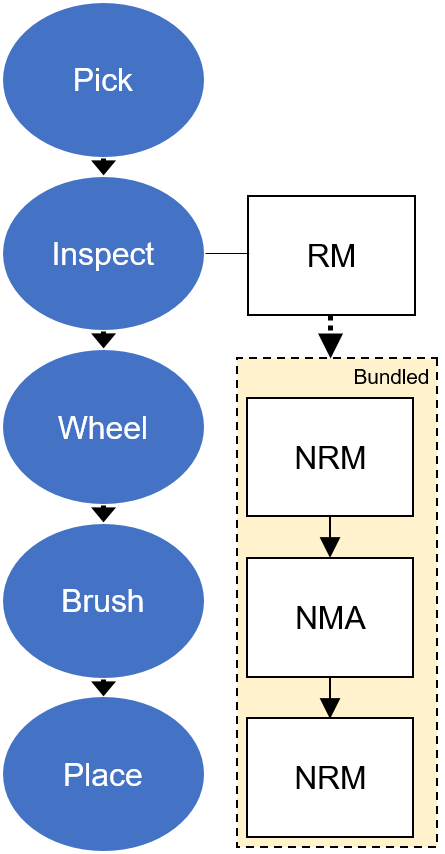}
	\caption{Decomposition of a single part deburring application into five sequential tasks shown in blue. Each task is decomposed into four sub-tasks as shown on the right. Some sub-tasks require the robot to be non-reactive (e.g. during object deburring), where the robot path/speed cannot change. These tasks are bundled and shown in the yellow box.}
	\label{deburringsteps}
\end{figure}

The workcell also consists of a single part feeder, two inspection cameras, a single material-removal ``wheel'', and two ``brushes'' for performing the deburring application. There is no external perception and static objects in the environment are modeled offline. The robot controllers exchange robot joint angle information, which is then used to generate robot occupancy information. In our test setup, the inspection cameras and the wheels are emulations. The center wheel and part feeder (for pick and place) are used by both robots, which creates potential goal overlap scenarios. 

\begin{figure}[h]
	\centering
	\includegraphics[scale =0.3]{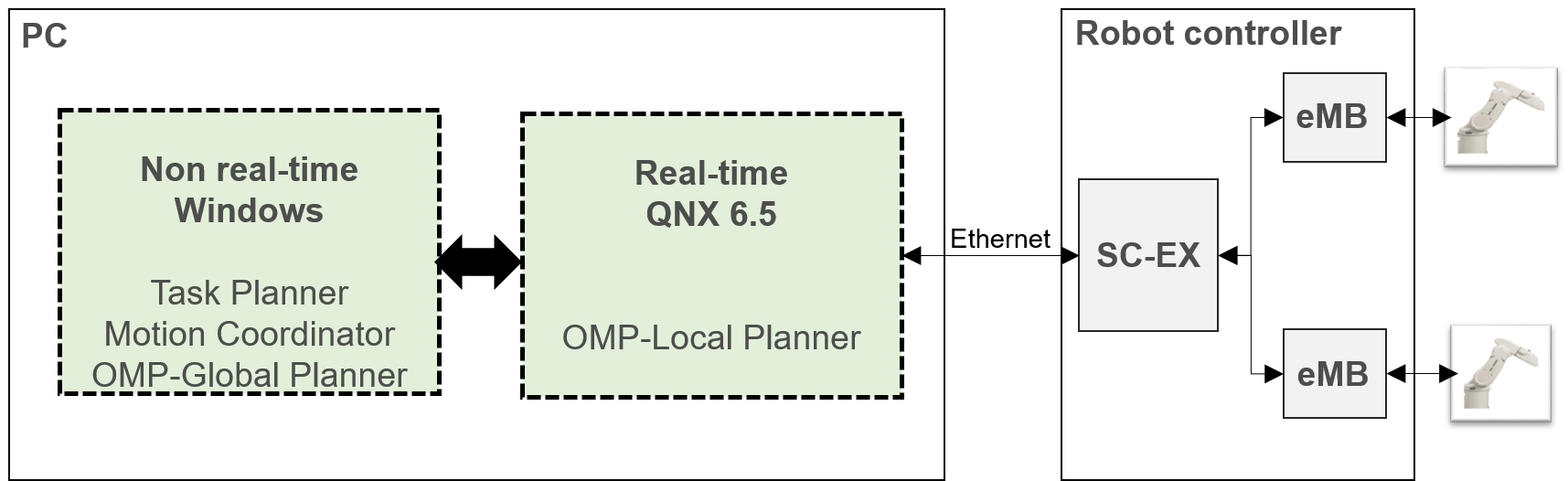}
	\caption{Hardware architecture for evaluation of the Multi-FLEX framework. The Task Planner, Motion Coordinator, and Global Planner are hosted on a non-real-time Windows system while the Local Planner is hosted on a QNX-based real-time system. }
	\label{HWInterface}
\end{figure}

The deburring application contains the following sequence of operations for each gear workpiece.  Each robot can process a gear at the same time that the other robot is processing a different gear, resulting in multi-robot, parallel processing of this sequence:
\begin{enumerate}
	\item Pick the workpiece from the feeder.
	\item Move the workpiece to the inspection area and identify the workpiece.
	\item Move to the deburring wheel and deburr the inner diameter.
	\item Move to the deburring brush and deburr the castle teeth.
	\item Move to and place the workpiece on the feeder.	
\end{enumerate}
Finally, each gear is of different size, so the processing time varies at the wheel and brush steps accordingly.
Each of these process steps is broken into tasks and sub-tasks per Section~\ref{TP}, as shown in Fig.~\ref{deburringsteps}. 

 The compute hardware architecture for the implemented system is described in Fig.~\ref{HWInterface}. The Multi-FLEX framework is hosted on an industrial PC running a type 1 hypervisor system with an Intel Xeon E-2276ME CPU, a non real-time Windows operating system and a QNX based real-time operating system. The desired joint poses are sent to the robot controller using Ethernet. In this work, an OMRON SmartControllerEX is used as the robot motion controller and an eMotionBlox-60R is used as the distributed servo controller. 
\subsection{Experiment results}
This deburring application demonstrates the performance and functionality of the Multi-FLEX framework. The functionality is evaluated based on task completion and collision avoidance. Performance is measured relative to a single robot workcell programmed using offline methods.  This “relative performance” is expected to be less than 200\% due to a) the potential for interference between the robots in the shared part of the workspace and b) reactive motion planning is often slower than non-reactive motions, since robot speeds are dynamically adjusted to ensure collision avoidance. Six parts are deburred, with each robot assigned an equal number of parts.

 All tasks were successfully completed in the application, with goal conflicts and overlaps handled by the Motion Coordinator.
\begin{figure}[h]
	\centering
	\includegraphics[width=0.45\textwidth]{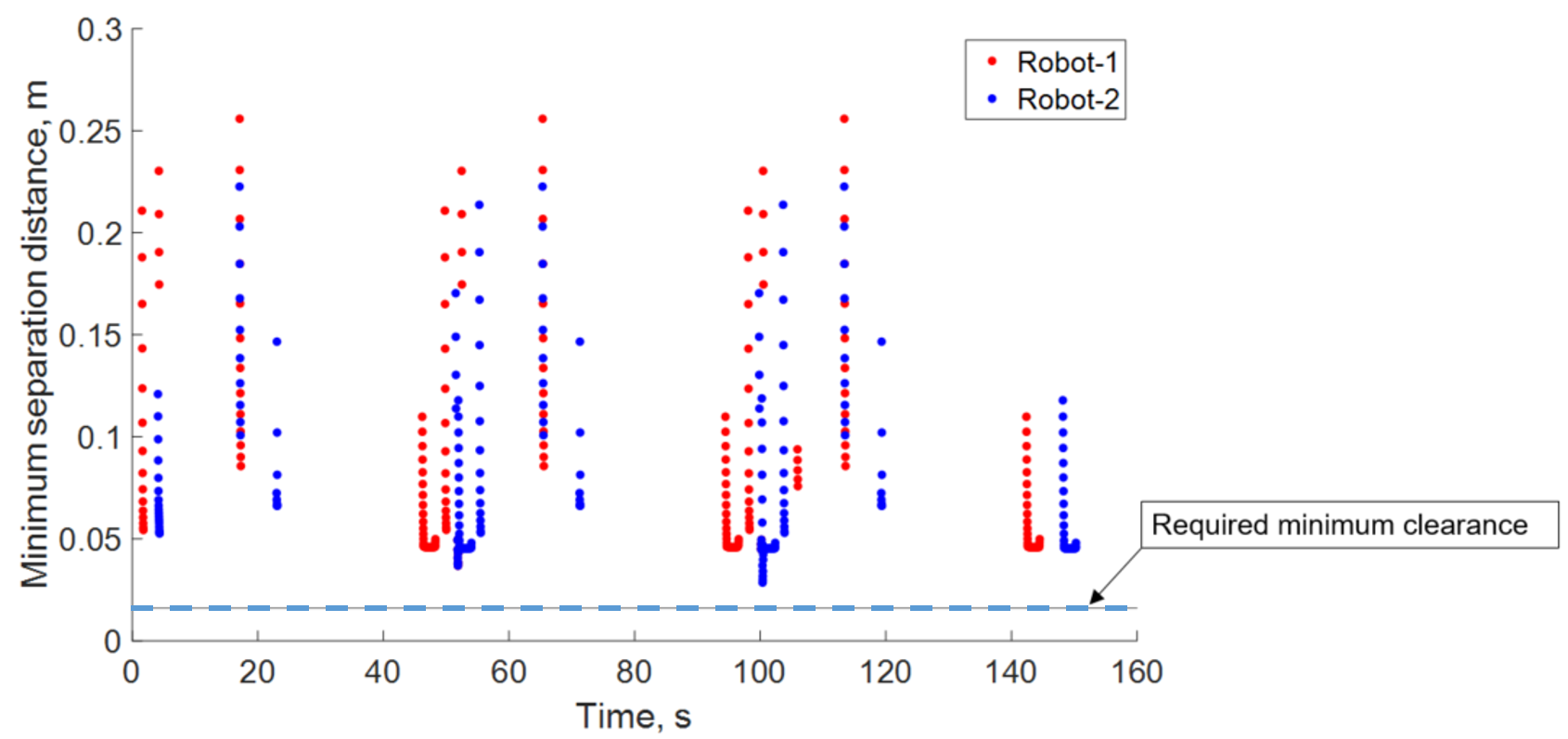}
	\caption{Robot-1 and Robot-2 minimum separation distance vs time for the robots during the deburring application. This shows that the minimum separation distances never violate the required minimum clearance.} 
	\label{collisiondistance}
\end{figure}

\begin{figure}[h]
	\centering 
	\subfloat[]{\includegraphics[width=4cm]{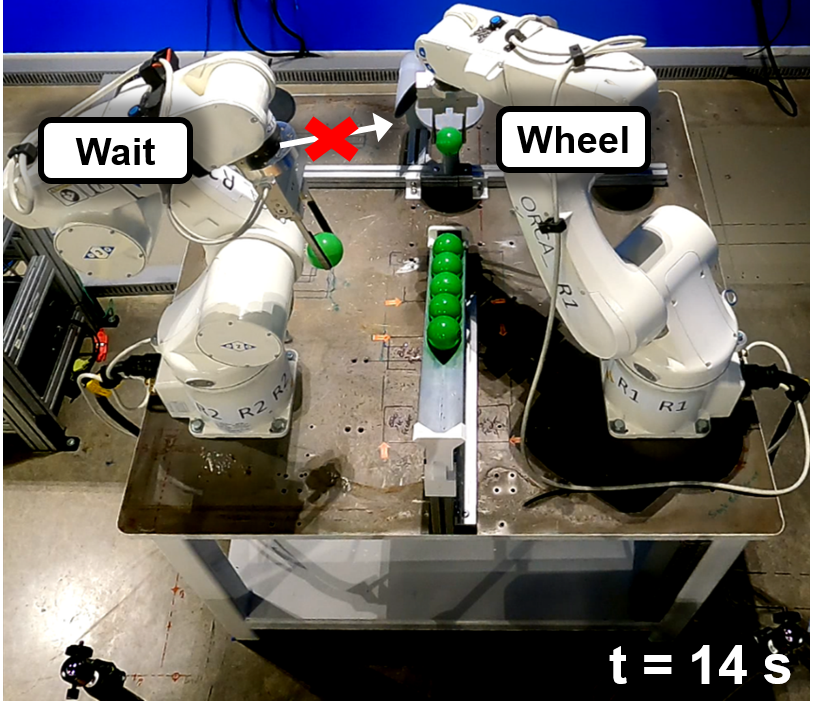}} 
	\quad 
	\subfloat[]{\includegraphics[width=4cm]{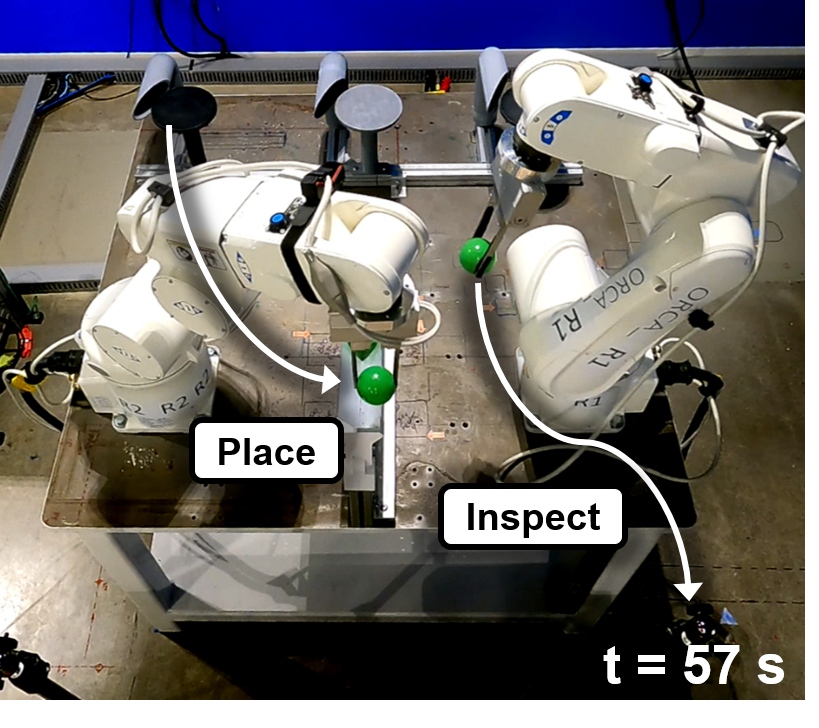}} 
	\caption{Snapshots of the deburring application with tasks labeled for each robot. In Figure (a), there is goal overlap at the wheel. The robot on the left is assigned a lower priority and the  automatically handles the overlap scenario by switching the state to `Wait' until the high priority task of the robot on the right is complete. In Figure (b), both robots are executing an RM sub-task, so the robots are reactive and avoid each other while completing the tasks without collision. A video synopsis of the Multi-FLEX enabled deburring application is available at https://www.youtube.com/watch?v=kZBThRgMmLs.}
	\label{snapshot}
\end{figure}

\begin{figure}[h!]
	\centering
	\includegraphics[width=0.5\textwidth]{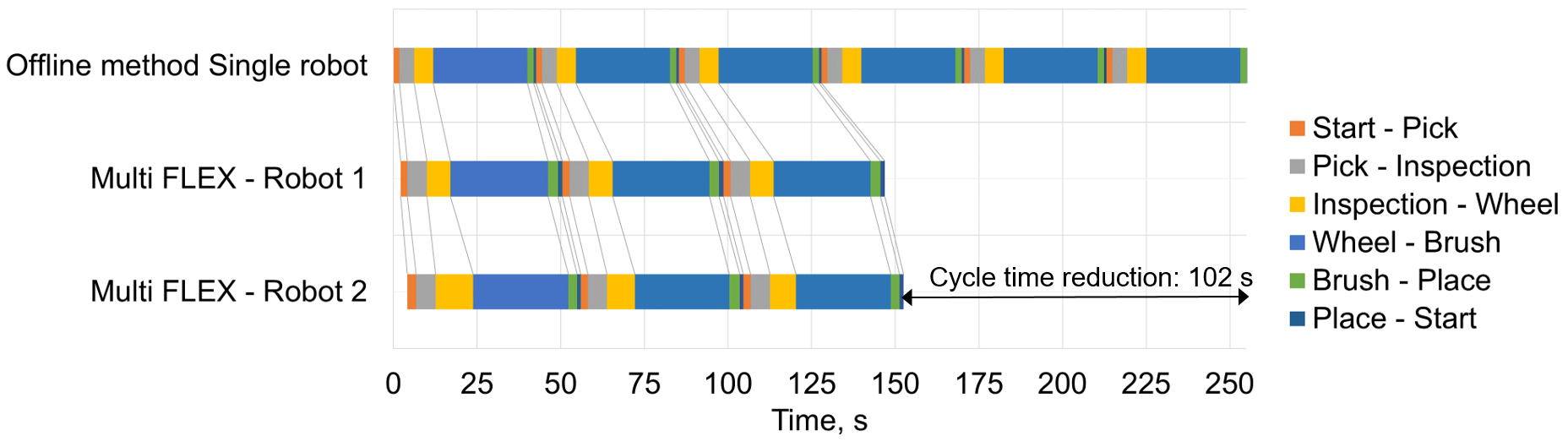}
	\caption{Cycle time savings for deburring six parts. This compares a two robot system using Multi-FLEX framework and a single robot station programmed using offline methods. The Multi-FLEX framework results in a higher performance application, as shown by the 102 s cycle time reduction. } 
	\label{performancechart}
\end{figure}

 The collision avoidance functional requirement was verified by evaluating the minimum distance between the robot and every other object in the environment (including the other robot). This minimum distance was checked against a required minimum clearance. The required minimum clearance is defined as $d_{eq}-\mu$, where $d_{eq}$ is the equilibrium distance and $\mu$ is a small positive number (to prevent numerical issues from triggering clearance violations). The minimum separation distances during the RM sub-tasks, when reactive collision avoidance is active, are shown in Fig.~\ref{collisiondistance}. 
Fig.~\ref{snapshot} illustrates two snapshots of robot motions during the deburring application. 
The two-robot setup performance was measured to be 170\% of the single robot workcell. Specifically, the cycle time for the application with the Multi-FLEX approach was approximately 151 s vs 255 s for the single robot workcell, and this is shown in Fig.~\ref{performancechart}. 

As discussed earlier, traditional offline programming approaches can also be difficult to setup, integrate, and validate (and change) and thus are “low flexibility”. During the integration phase, the user typically a) decomposes the application into individual tasks, b) selects appropriate robot poses for each task, and c) programs individual motions for the task to ensure collision avoidance and task success. This last step, programming individual motions, is often iterative and can be very time-consuming. Multi-FLEX makes this step simple by automatically determining individual robot motions and handling collision avoidance. This results in a solution that is easy to setup and change, and thus is “high flexibility”.

\section{CONCLUSIONS}
This letter presented an integrated task and reactive motion planning framework called Multi-FLEX that is well-suited for real-world, industrial multi-robot applications. To fully realize the benefits of reactive motion planning in such applications, it is necessary to incorporate industrial application constraints. Multi-FLEX enables this by using the concepts of task dependency accommodation, task decomposition, and task bundling.

The proposed framework is evaluated on a deburring use case with high-speed industrial robots. The functional objectives, task completion and collision avoidance, and the performance objective, relative productivity, were achieved. Further, compared with difficult-to-use and low flexibility traditional approaches, Multi-FLEX is an easy-to-program and flexible-to-change solution. As a result, the Multi-FLEX framework provides a clear path for more widespread implementation of industrial multi-robot applications. 
\bibliographystyle{IEEETran}
\bibliography{ExportedItems.bib}
\end{document}